\documentclass[graybox]{svmult}

\usepackage{mathptmx}       
\usepackage{helvet}         
\usepackage{courier}        
\usepackage{type1cm}        
\usepackage{makeidx}         
\usepackage{graphicx}        
\usepackage{multicol}        
\usepackage[bottom]{footmisc}
\usepackage{booktabs} 

\makeindex             

\begin{document}

\title*{\centering Leveraging Chat-Based Large Vision Language Models for Multimodal Out-Of-Context Detection}
\titlerunning{Leveraging Chat-Based LVLMs for Multimodal OOC Detection}

\author{\centering Fatma Shalabi$^{*1,2}$,  Hichem Felouat$^{*1,2}$, Huy H. Nguyen$^{2}$, and Isao Echizen$^{1,2,3}$\\
\small{$^{*}$These authors contributed equally} \\
\small{$^{1}$The Graduate University for Advanced Studies, SOKENDAI, Japan} \\
\small{$^{2}$National Institute of Informatics, Japan \ \ \ \ \ \ \ \ \ \ \ \ \ \ \ \ \ \ \ \ \ \ \ \ \ \ \ $^{3}$The University of Tokyo, Japan} \\
\small{E-mail: \{fatmafaek, hichemfel,  nhhuy, iechizen\}@nii.ac.jp}}
\authorrunning{Fatma Shalabi \textit{et al.}}
\maketitle

\abstract{Out-of-context (OOC)  detection is a challenging task involving identifying images and texts that are irrelevant to the context in which they are presented. Large vision-language models (LVLMs) are effective at various tasks, including image classification and text generation. However, the extent of their proficiency in multimodal OOC detection tasks is unclear. In this paper, we investigate the ability of LVLMs to detect multimodal OOC and show that these models cannot achieve high accuracy on OOC detection tasks without fine-tuning. However, we demonstrate that fine-tuning LVLMs on multimodal OOC datasets can further improve their OOC detection accuracy. To evaluate the performance of LVLMs on OOC detection tasks, we fine-tune MiniGPT-4 on the NewsCLIPpings dataset, a large dataset of multimodal OOC. Our results show that fine-tuning MiniGPT-4 on the NewsCLIPpings dataset significantly improves the OOC detection accuracy in this dataset. This suggests that fine-tuning can significantly improve the performance of LVLMs on OOC detection tasks.
}

\section{Introduction}
\begin{figure}[ht]
     \begin{center}
     \includegraphics[scale=0.35]{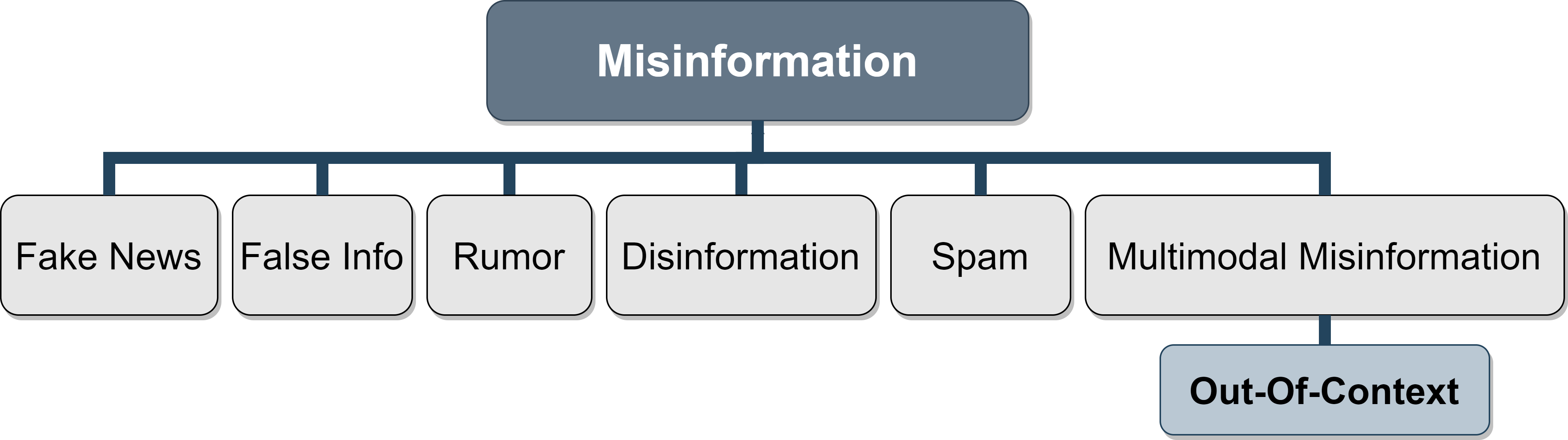}
     \caption{Key types related to misinformation ~\cite{islam2020deep}.}
     \label{fig:types}
     \end{center}
\end{figure}

Since misinformation is growing rapidly in digital communication channels, it refers to false or misleading information that spreads rapidly through these channels, including social media, news articles, and word-of-mouth. It can have harmful consequences, such as leading people to make poor decisions regarding their health, finances, and safety. Multimodal misinformation is a particularly concerning form of misinformation that combines images with text to deceive or mislead people, making it appear more realistic and challenging to detect. Many terms are related to misinformation, such as rumor, fake news, false information, spam, disinformation, and multimodal misinformation ~\cite{lin2019rumor} ~\cite{islam2020deep}. Figure \ref{fig:types} illustrates the related terms of misinformation.

\begin{figure}[ht]
     \begin{center}
     \includegraphics[scale=0.34]{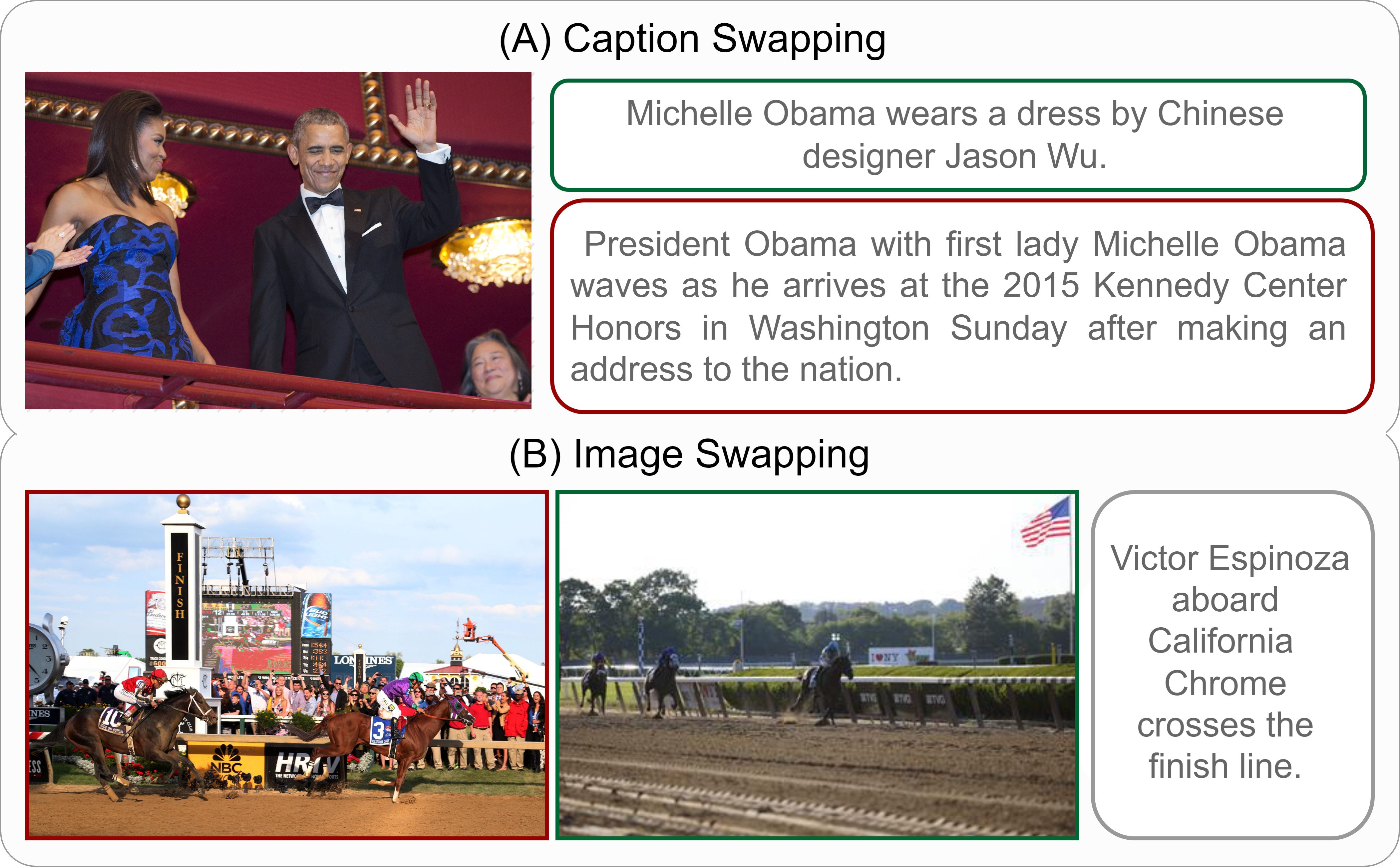}
     \caption{Examples from our dataset show that OOC content generation arises from swapping authentic images and captions. In example A, the red caption, taken from a different context, is paired with the image. In example B, the green image is the original visual representation of the caption, while the red image is from a different context.
     }
     \label{fig:out_of_context}
     \end{center}
\end{figure}

One common type of multimodal misinformation is out-of-context (OOC), which involves separating authentic images from their original context and pairing them with misleading texts. This can lead to the loss or change of the intended meaning and can be used to deceive or mislead the audience intentionally. Figure~\ref{fig:out_of_context} depicts an example of a multimodal OOC concept, where authentic images and captions are mismatched to create a deceptive narrative. This manipulation intentionally alters the original message, potentially misleading or deceiving the audience. To address this problem, researchers are developing systems that can detect OOC information in images and texts. One approach uses pre-trained VLMs, which have demonstrated effectiveness across various tasks. Fine-tuning the parameters of these models enables them to be adapted to specific downstream tasks, such as OOC detection. Fine-tuning LVLMs involves training the models on a labeled OOC dataset and authentic image-text pairs. This process enables the models to learn the nuances of OOC manipulation and identify subtle inconsistencies between images and texts that may indicate OOC content ~\cite{su2019vl} ~\cite{li2019visualbert}.

\begin{figure}[ht]
     \begin{center}
     \includegraphics[scale=0.17]{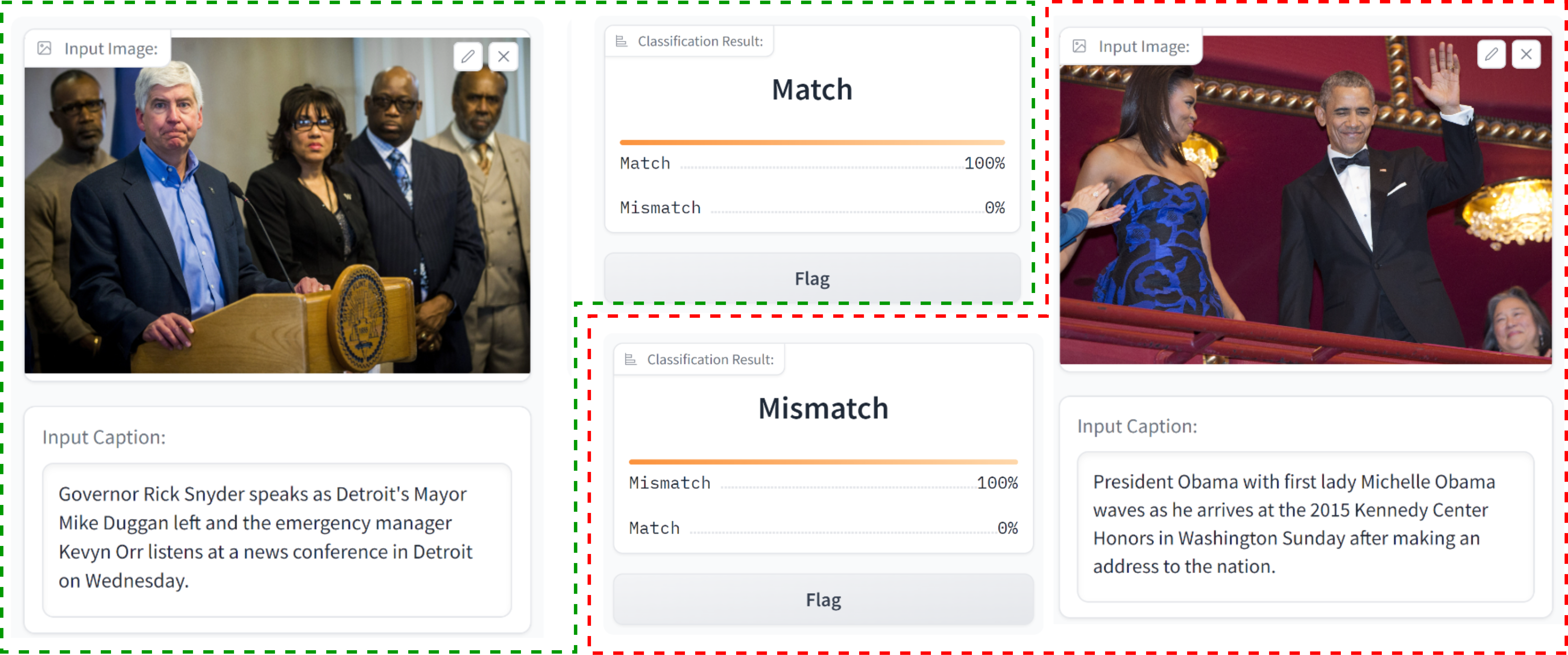}
     \caption{Illustration of our model's ability to detect contextual consistency in an image caption. Green: The image and caption are in context (Match). Red: The image and caption are out of context (Mismatch).
     }
     \label{fig:our_contribution}
     \end{center}
\end{figure}

LVLMs are well-suited for OOC detection tasks due to their ability to comprehend the relationship between visual and textual information. This means they can identify anomalies in a scene, even if they are not explicitly mentioned in the text. Also, LVLMs can learn from various data sources, enabling them to detect various OOC situations. Moreover, LVLMs can be fine-tuned for specific tasks, making them adaptable to detecting types of anomalies. Inspired by LVLMs' success in computer vision, this study comprehensively evaluated the OOC detection capabilities of MiniGPT-4 on a  well-designed dataset. Figure~\ref{fig:our_contribution} shows the model's effectiveness in pinpointing multimodal anomalies, highlighting its potential for this crucial task. 

The key contributions of this study encompass the following: We investigated the ability of LVLMs to detect OOC content. Our findings demonstrate that these models cannot achieve high accuracy on OOC detection tasks. We demonstrate that fine-tuning LVLMs on OOC datasets can further improve their OOC detection accuracy. This suggests that fine-tuning can help these models learn more specific features that are indicative of OOC content. The rest of this paper is structured as follows: Section 2 surveys existing techniques for OOC detection. Section 3 then details our proposed approach. Section 4 presents the results and analysis of our experiments, followed by a discussion of the findings and avenues for future work. Concluding remarks are presented in Section 5.

\section{Related Work}
\subsection{Vision Language Models}
Advancements in machine learning have given rise to powerful techniques for pre-training visual-language representations. The CLIP model~\cite{radford2021learning}, a prominent neural network architecture, leverages the inherent semantic connections between language and vision to extract transferable visual representations from natural language descriptions. Trained on a massive dataset of (image, text) pairs, CLIP effectively predicts a given image's most relevant textual description. This capability, similar to the zero-shot learning abilities of GPT-2~\cite{radford2019language} and GPT-3~\cite{brown2020language}, facilitates effective cross-modal understanding.

Leveraging pre-trained language models has demonstrated efficacy in enhancing the comprehension of both visual and textual information using machines. Flamingo~\cite{alayrac2022flamingo} further improved this idea by aligning a pre-trained image processor and language model through gated cross-attention. This model learned from billions of image-text pairs and could quickly learn from a small number of examples. BLIP-2~\cite{li2023blip} built on this using Flan-T5~\cite{chung2022scaling} and Q-Former to better connect visual features with language.

The release of GPT-4~\cite{openai2023gpt4} has brought even better visual understanding and reasoning abilities. Models like ChatGPT, which is a type of language model, have been effective in working together with other specialized models for vision and language tasks. For example, Visual ChatGPT~\cite{wu2023visual} and MM-REACT~\cite{yang2023mm} have shown how ChatGPT can collaborate with different visual models to handle more complex tasks. On the other hand, MiniGPT4~\cite{zhu2023minigpt} directly connects visual information with the language model to accomplish different tasks involving images and texts without needing external vision models.

\subsection{Multimodal Misinformation Detection}
Misinformation is a growing threat, as using a combination of text with images is a valid tool for creating realistic-looking misinformation called multimodal misinformation, often leading to real-world consequences like public distrust, political instability, and so on. Researchers have proposed several methods for detecting multimodal misinformation to address this threat. One approach is extracting valuable features from images and texts and then aligning them before feeding them into a classifier. For example, Singhal \textit{et al.} ~\cite{singhal2022leveraging} proposed an approach that suppresses information from weaker modalities and extracts relevant information from the strongest modality. They used a gating mechanism to control the flow of information between the text and image modalities and were trained to determine which modality was more informative for a given sample.

Another approach focuses on fusing the multimodal features using a co-attention mechanism. For example, Wu \textit{et al.}~\cite{wu2021multimodal} proposed the use of stacking multiple co-attention layers to learn the relationships between the text and image features. Jing \textit{et al.}~\cite{jing2023multimodal} proposed a progressive fusion network (MPFN) for multimodal disinformation detection. The MPFN captures the representational information of each modality at different levels and progressively fuses the information from the same level and different levels using a mixer to establish a strong connection between the modalities. Zhang \textit{et al.}~\cite{zhang2023detecting} relied on neural-symbolic multimodal learning methods to propose a model that symbolically disassembles the text-modality information to a set of fact queries based on the abstract meaning representation of the caption and then forwards the query-image pairs into a pre-trained LVLM to select the ``evidence'' that enables them to detect misinformation.

Moholdt \textit{et al.}~\cite{moholdt2023detecting} developed a method to detect OOC by comparing AI-generated images and captions from the COSMOS dataset~\cite{aneja2023cosmos}. They created new datasets using synthetic text-to-image generative models, Stable Diffusion, and DALL-E 2, used object detection and encoding to compute image similarity, and employed cosine similarity for comparison.

\subsection{Leveraging Pre-trained Vision Language Models in Multimodal OOC Detection}
In recent years, the field of multimodal OOC detection has seen significant advancements, driven in part by the emergence of powerful pre-trained VLMs. Leveraging these models has become a key research focus, as they offer the potential to improve the accuracy and robustness of OOC detection systems. In this section, we review related work that explores the integration of pre-trained VLMs into the context of OOC detection. Luo \textit{et al.}~\cite{luo2021newsclippings} developed a new method for identifying mismatches between images and their corresponding captions. Their method uses the large pre-trained VLM, CLIP,  to classify mismatches based on retrieval. They also evaluated both CLIP and VisualBERT ~\cite{li2019visualbert} on the proposed dataset and achieved classification accuracies of 60.23\% and 54.81\%, respectively. However, they also found that both machine and human mismatch detection is still limited, suggesting that this task is challenging.

Huang \textit{et al.}~\cite{huang2022text} used CLIP and VinVL~\cite{zhang2021vinvl} to detect inconsistencies in multimedia content. They encoded each image and its corresponding caption separately and then compared their embeddings. Dissimilar embeddings suggested that the text-image pair was out of context. The authors evaluated their method on several large-scale datasets. However, their method does not address the potential biases in the VLMs used, which may have affected the accuracy of the OOC detection task. Additionally, their method relies on the availability of textual information associated with the image, which may not always be available.

Fatma \textit{et al.}~\cite{shalabi2023imagetext} presented a novel methodology for detecting OOC content by employing synthetic data generation. They used two large pre-trained VLMs: BLIP-2 to generate captions describing original images and Stable Diffusion to generate images from original captions. Their approach calculated the similarity between original and generated data using CLIP, Vision transformer (ViT)~\cite{dosovitskiy2021image}, and Sentence-BERT~\cite{reimers2019sentence} to assess the coherence of the input image-caption pair. This research validated the efficacy of synthetic data generation in addressing data limitations associated with OOC detection, achieving a classification accuracy of 68.0\%.

\section{Proposed Approach}
We propose a new method for OOC detection that relies on LVLMs, as shown in Figure ~\ref{fig:our_approach}. Our method includes MiniGPT-4 ~\cite{zhu2023minigpt}, a new VLM that can generate detailed image descriptions, construct websites from handwritten drafts, and craft stories and poems inspired by images.

The model undergoes a two-stage training process. In the first stage, the model learns the basics of visual and linguistic domains by training on a large collection of raw image-text pairs. It uses a ViT backbone fused with a querying transformer inspired by BLIP-2 to process images. However, this initial training phase produces unnatural language output, leading to the need for a second stage focused on improving natural language generation.

The second stage involves training the model on a curated dataset structured like a conversation, with questions and corresponding answers. A linear projection layer ensures seamless alignment between extracted visual and linguistic information. This layer alone is trained, while the pre-trained vision encoder and language decoder remain frozen. The layer's output feeds into Vicuna ~\cite{chiang2023vicuna}, a fine-tuned open-source chatbot that generates natural language descriptions based on fine-tuning LLaMA ~\cite{touvron2023llama}. This stage enhances the association between images and text, producing more reliable and natural language outputs.

\subsection{Fine-tuning Large Language Models}
Fine-tuning has been widely used in recent large language model (LLM)  studies to align model outputs with specific task or domain requirements ~\cite{ji2023survey}. Without fine-tuning, LLMs are susceptible to biases, inaccuracies, and irrelevance to the context of a particular application. Fine-tuning effectively enriches LLMs with domain knowledge, enhances their task-specific capabilities, improves the fairness and reliability of their outputs, and mitigates potential harm caused by hallucinations ~\cite{ji2023survey}. However, fine-tuning LLMs incurs significant computational costs and demands substantial domain-specific data, which may not be readily available due to privacy concerns. Hu \textit{et al.}~\cite{hu2021lora} proposed adapting pre-trained LLMs to specific domains by fine-tuning modules with limited trainable parameters to address the computational challenge. To address the data privacy concern, federated learning (FL) emerged as a promising solution where a distributed learning paradigm enables multiple entities to optimize a model collaboratively without directly sharing their data ~\cite{kuang2023federatedscope}.

\subsection{OOC Problem Formulation}
In general, the formulation of the OOC detection task focuses on the samples that contain both text and image. Let each sample be \(X = (x_{img}, x_{txt})\). Denote the ground-truth label as \(Y\), when \(Y = 0\), \(X\) is a match; otherwise, it is a mismatch. A rich set of features are first extracted from \(x_{img}\) and \(x_{txt}\), then fused and projected into a single value of \({y}'\), i.e., match or mismatch.

\begin{equation}
     \label{eqn:1}
     {y}' = F_{cls}(F_{Mix}(F_{img}(x_{img}),F_{txt}(x_{txt})))
\end{equation}

The procedure is depicted in Equation \ref{eqn:1}, where \(F_{img}\) and \(F_{txt}\) are unimodal feature extractors, \(F_{Mix}\) is a feature fusing model, and \(F_{cls}\) is the classification head. Instead of applying sophisticated and black-box feature fusing networks, we propose a simple yet effective method that relies on leveraging the VLM.

\subsection{Fine-tuning MiniGPT-4}
To fine-tune MiniGPT-4 for the specific task of OOC detection, the first essential requirement pertains to assembling a dataset comprising \textit{n} \((x_{img}, x_{txt})\) image-caption pairs. Additionally, these pairs must be labeled \textit{Y} as either match or mismatch pairs. Subsequently, a transformative step involves restructuring this dataset to match the structure of MiniGPT-4's dataset, on which it has been trained in the second stage. This transformative procedure establishes a harmonious alignment between the task-specific dataset and the existing knowledge within MiniGPT-4, thereby facilitating the subsequent fine-tuning process. Moreover, as an auxiliary step, we need to adjust Vicuna's output to match the actual \textit{True label} format during training. This change enables us to compare Vicuna's output \({y}'\) (predicted label) with the actual \textit{True label} \textit{Y}, which aids in calculating the loss and accuracy during each epoch.

\subsubsection{Data Preparation}
In the second stage of MiniGPT-4 training, the authors used a dataset of image-caption pairs. The prompts were used to guide MiniGPT-4 in generating descriptions for the images. The generated descriptions were then evaluated against the original captions to measure accuracy. To improve the accuracy and usability of our multimodal detection model, we fine-tuned MiniGPT-4 on a labeled multimodal dataset. We modified this dataset structure to match the model's input after our modifications, resulting in the following format: \textit{(image, caption, label)}. The model's expected output was a simple \textit{``Yes'' or ``No''} response. The loss was computed by comparing the model's output with the true label assigned to the pair.

\subsubsection{Fine-tuning MiniGPT-4 for Multimodal OOC Detection}

\begin{figure}[ht]
     \begin{center}
     \includegraphics[scale=0.31]{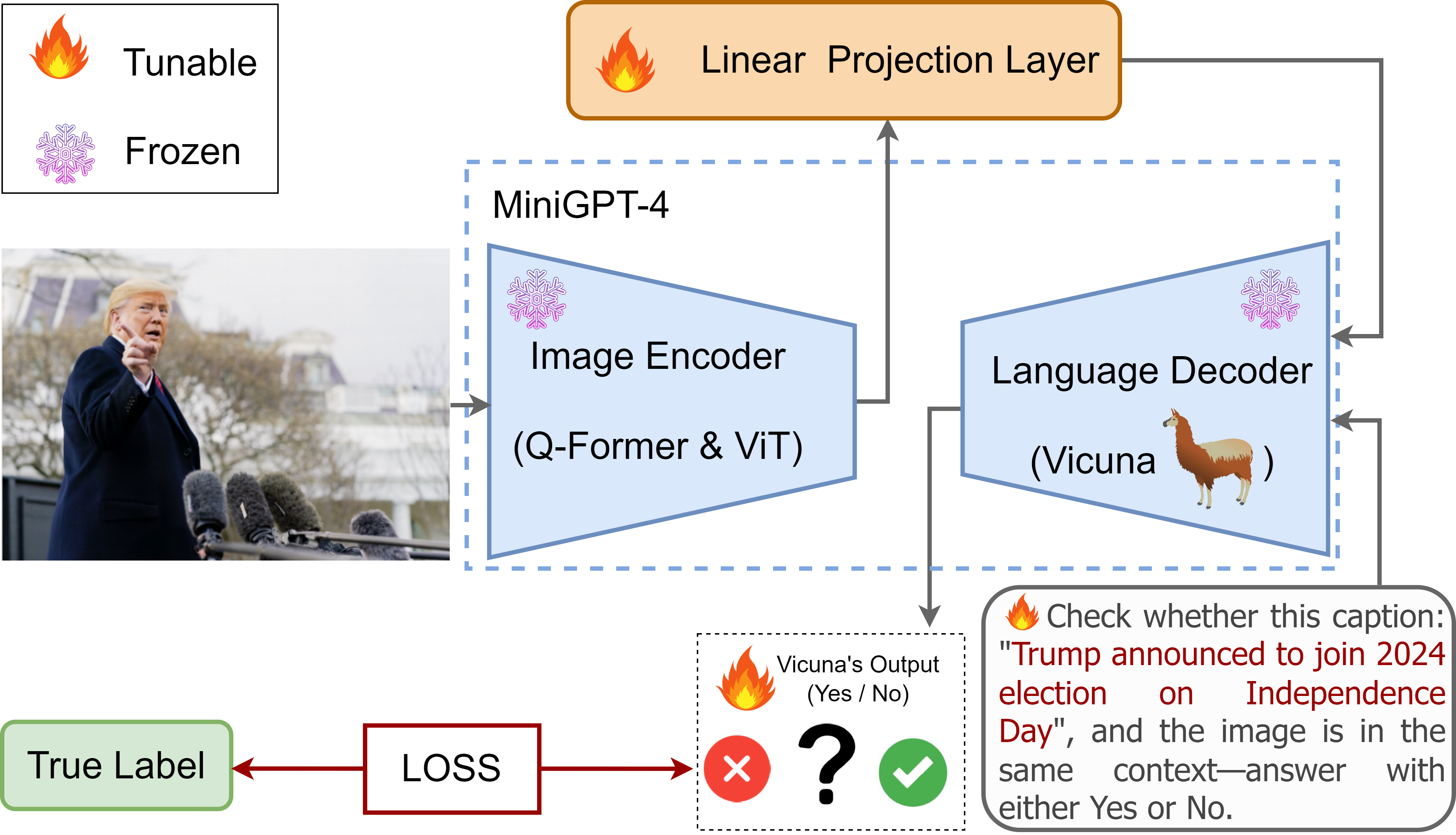}
     \caption{The workflow of our approach entails exclusively training the linear projection layer to align visual features with the Vicuna and establish weights consistent with our dataset structure. Throughout the training phase, we directed MiniGPT-4 to generate responses in a binary \textit{``Yes'' or ``No''} format to verify the contextual relevance of the image caption relative to the provided image.
     }
     \label{fig:our_approach}
     \end{center}
\end{figure}

The original MiniGPT-4 model is trained on triplets of images, prompts, and corresponding image descriptions. During training, it takes an image $\textit{(img)}$ and a prompt $\textit{(p)}$ as input and attempts to generate a new description $d_{pred}$ that closely matches the original description provided $d_{true}$. This is achieved by minimizing a loss function that measures the difference between the generated and original descriptions, as shown in Equation \ref{eqn:2} on the left side. When used for inference, the model only requires an image and prompt as input and outputs a predicted description of the image, as shown in  Equation \ref{eqn:2} on the right side.

\begin{equation}
     \label{eqn:2}
      M_{train}(img, p, d_{true}) \rightarrow d_{pred} \hspace{0.25cm}|\hspace{0.25cm}  M_{inf}(img, p) \rightarrow d_{pred}
\end{equation}

In our approach, we fine-tuned the MiniGPT-4 model for the multimodal OOC detection task. This involved providing both an image $\textit{(img)}$ and prompt $\textit{(p)}$ to the model, where the prompt consisted of a combined question and caption $\textit{(f(q, cap))}$, a function $\textit{f}$ used to combine the question and caption, and attempts to predict a label $lab_{pred}$, as shown in  Equation \ref{eqn:3} on the right side. We used the weights from the first training stage to align the model with the specific requirements of our dataset's modified structure. Subsequently, we focused on retraining the second stage using this approach to acquire weights compatible with the modified dataset structure. 

\begin{equation}
     \label{eqn:3}
     {M}'_{train}(img, f(q, cap), lab_{true}) \rightarrow lab_{pred} \hspace{0.25cm}|\hspace{0.25cm} {M}'_{inf}(img, f(q, cap)) \rightarrow lab_{pred}
\end{equation}

During training (Equation \ref{eqn:3} on the left side), the fine-tuned MiniGPT-4 model takes an image and a combined prompt consisting of a question and a caption. Its objective is to predict a label $lab_{pred}$ matching the true label $lab_{true}$. Since the nature of this task is a binary classification, we used the CrossEntropy loss function (Equation \ref{eqn:4}).

\begin{equation}
     \label{eqn:4}
     L(x, y) = \left \{ l_{1},...,l_{n} \right \}^{T} = l_{n} = -w_{yn} \times log\left( \frac{exp(x_{n,yn})}{\sum_{c=1}^{2} exp(x_{n,c}) } \right ) 
\end{equation}

This function guides the fine-tuned MiniGPT-4 towards generating binary responses \textit{("Yes/No" or "Match/Mismatch")} while also adjusting to be compatible with the new input format of the combined prompt (Equation \ref{eqn:5}).  This approach enabled us to calculate the loss by comparing the model’s output with the ground-truth labels. Figure ~\ref{fig:our_approach} shows a flowchart illustrating these key steps in our approach.

Where \textit{x} is the input, \textit{y} is the target, \textit{w} is the weight, and \textit{N} is the batch size.
\begin{equation}
     \label{eqn:5}
     {L}'(x = (img, f(q, cap)), y = lab_{true})   
\end{equation}

\section{Experimental and Results}
\subsection{Dataset}
We used the NewsCLIPpings dataset ~\cite{luo2021newsclippings}, which is a large dataset of challenging mismatched image-caption pairs constructed based on the VisualNews corpus ~\cite{liu2021visual}. It includes news articles from four major news outlets: the BBC, The Guardian, The Washington Post, and USA Today. The dataset was created using various techniques to introduce mismatches between images and captions, such as using semantic similarity between images and captions, matching captions semantically but with different images, associating captions mentioning the same individuals, and identifying captions describing similar scenes. The dataset provides samples representing challenging mismatches between captions and images that can mislead humans. Table 1 presents several statistics for the NewsCLIPpings dataset.

\subsection{Experimental Settings}
The experiments were conducted on an NVIDIA A100 G80 GPU. The model was trained with a batch size of 4 for a total of 30 epochs. The number of iterations per epoch was equal to the number of examples in the dataset divided by the batch size.

\subsection{Results and Discussion}
\begin{figure}[h]
     \center
     \includegraphics[scale=0.43]{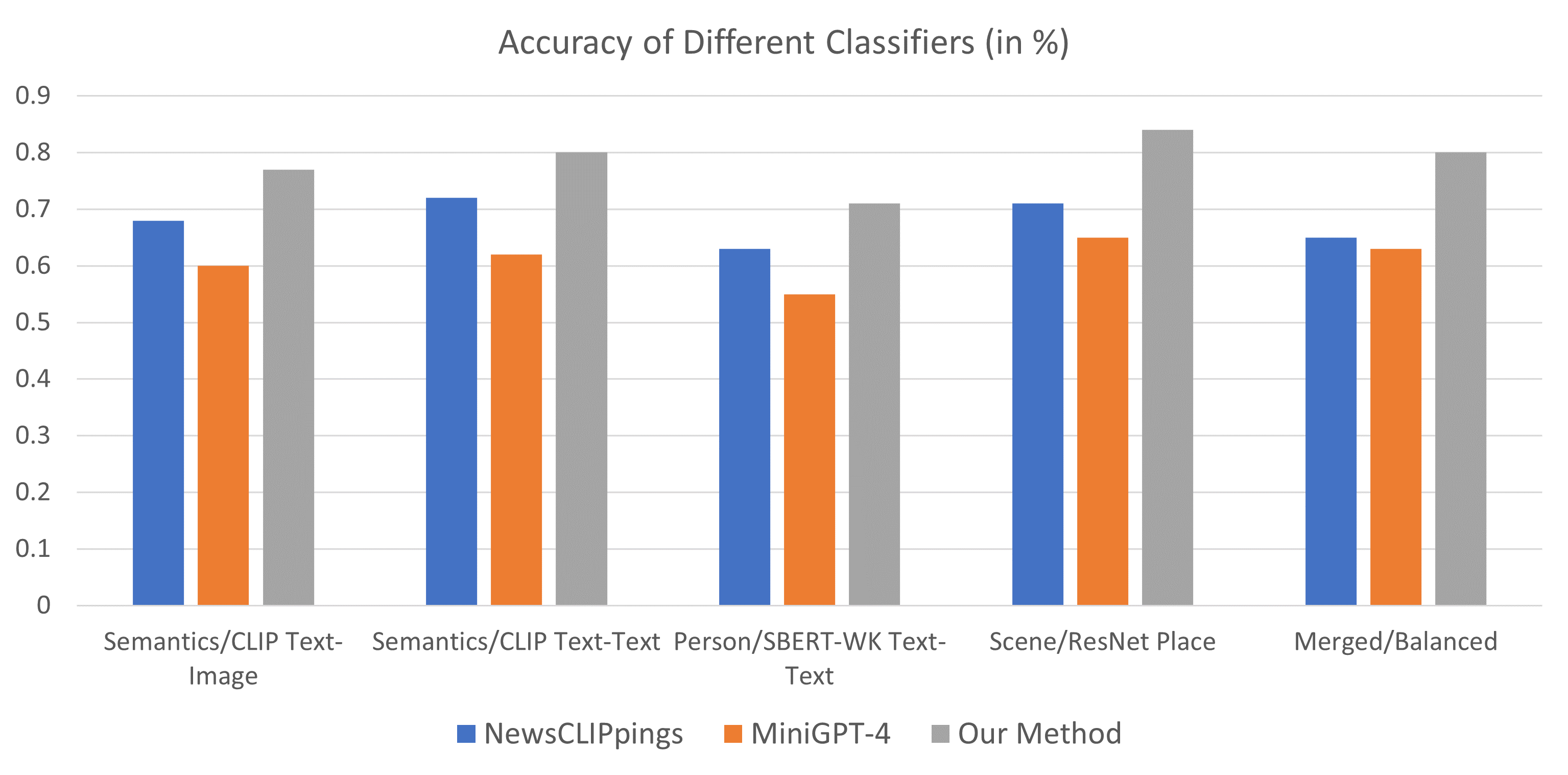}
     \caption{Our method achieves accuracy gains of $\geq 8\%$ across diverse classification splits of the NewsCLIPpings dataset, compared to NewsCLIPpings and MiniGPT-4 (zero-shot) classifiers.
     }
     \label{fig:MiniGPT4_result_comp}
\end{figure}

We present our results in Table~\ref{tab:comparison_result_NewsCLIPpings} and compare them with the best results obtained in the NewsCLIPpings paper for each dataset split ~\cite{luo2021newsclippings}. Our method outperforms the NewsCLIPpings paper results on every dataset split with a gain of $\geq 8\%$ as shown in Figure ~\ref{fig:MiniGPT4_result_comp}, indicating that fine-tuning LVLMs provides fundamental improvements in OOC detection. We further validate our results by comparing them with other methods in the same field that used the Merged/Balanced split of the NewsCLIPpings dataset, excluding methods that rely on additional information from the Internet for classification. As shown in Table ~\ref{tab:comparison_result_other_methods}, our approach is among the best methods for OOC detection.

\begin{figure}[t]
     \center
     \includegraphics[scale=0.3]{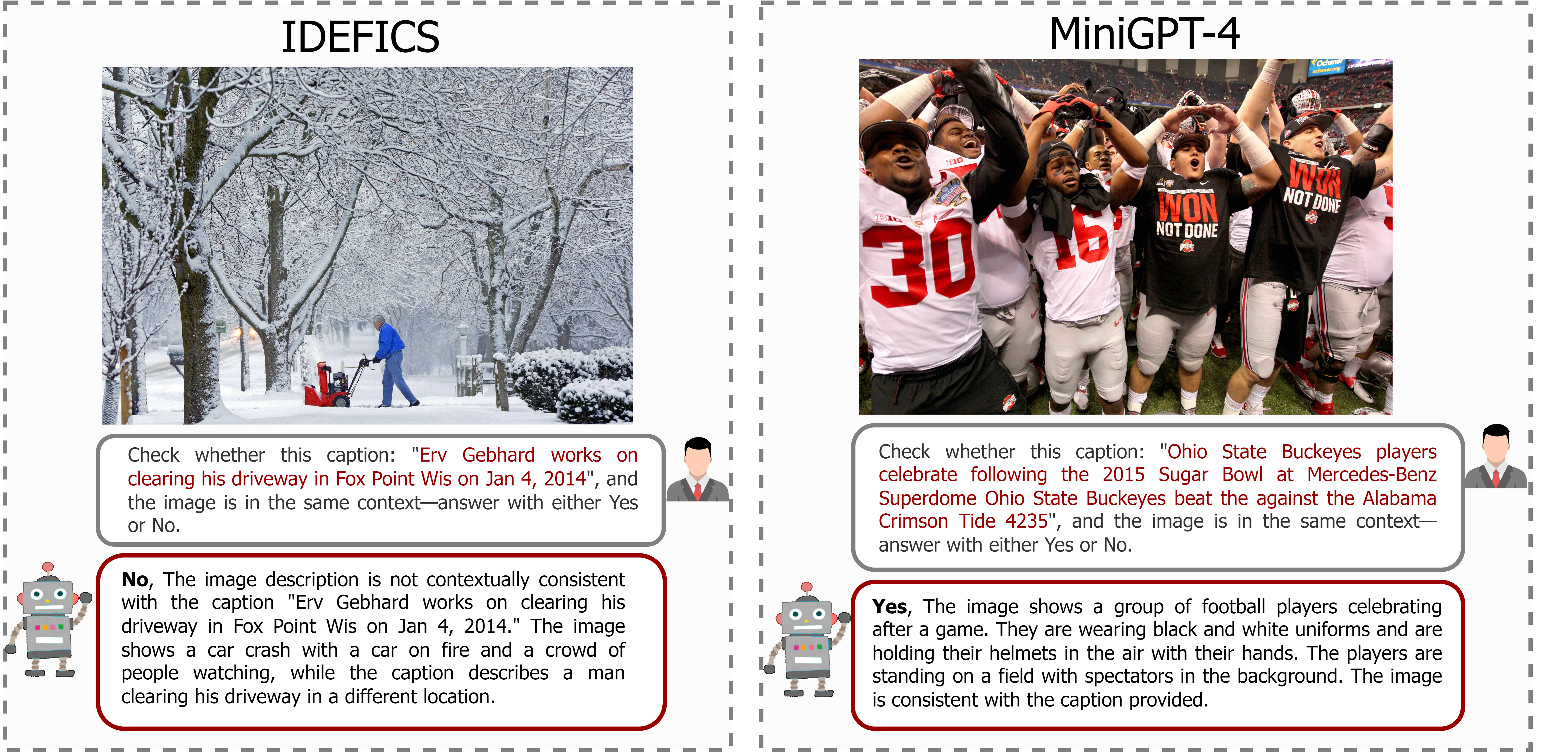}
     \caption{OOC detection results for LVLMs MiniGPT-4 and IDEFICS without fine-tuning.
     }
     \label{fig:vlm_without_fine_tuning}
\end{figure}

In our experiments with LVLMs for OOC detection (MiniGPT-4 ~\cite{zhu2023minigpt} and IDEFICS\footnote{https://huggingface.co/blog/idefics}), as shown in Figure ~\ref{fig:vlm_without_fine_tuning}, we observed the following limitations: LVLMs tend to provide descriptive responses rather than direct answers, making it challenging to evaluate their accuracy. The lack of a clear evaluation metric hinders our ability to determine the accuracy of LVLM-generated descriptions. These limitations suggest that LVLMs may not be the most suitable tool for OOC detection tasks.

However, we used MiniGPT-4 to assess the contextual coherence between image-caption pairs. We prompted the model with questions requiring \textit{(``Yes'' or ``No'')} answers concerning the alignment between the image and its caption. Nevertheless, instead of directly providing \textit{Yes/No} responses, the model generated descriptions that incorporated the intended response, as shown in Figure ~\ref{fig:vlm_without_fine_tuning}. We developed a post-processing step to evaluate performance, extracting the core \textit{Yes/No} answer from these descriptions. This enabled us to obtain binary classification results, which were used to calculate MiniGPT-4's performance in zero-shot inference. These results are presented in Table ~\ref{tab:comparison_result_NewsCLIPpings} and Figure ~\ref{fig:MiniGPT4_result_comp}.

Fine-tuning LVLMs for OOC detection enables them to produce direct binary responses \textit{(``Yes'' or ``No'')}. This approach facilitates a more straightforward assessment and comparison of their performance compared to relying on descriptive responses. While this approach has shown promise in this field, it falls short in providing explanations for the answers it generates. This is an area we intend to address in future research as we strive to enhance this approach into a framework capable of both classifying and interpreting the responses it provides.
\begin{table}[t]
    \center
     \caption{Comparison between our performance results, NewsCLIPpings paper results  ~\cite{luo2021newsclippings} (Accuracy, Pristine, Falsified, and AUC in \%, and MiniGPT-4: Zero-Shot inference of MiniGPT-4).}
     \label{tab:comparison_result_NewsCLIPpings}
     \resizebox{12cm}{!}{%
    \begin{tabular}{llll|lll|lll|llll}
    \hline
     &
     &
     &
     &
     \multicolumn{3}{c|}{\textbf{NewsCLIPpings}} &
     \multicolumn{3}{c|}{\textbf{MiniGPT-4}} &
     \multicolumn{4}{c}{\textbf{Our Method}} \\ \cline{5-14} 
    \textbf{Split} &
     \textbf{Train} &
     \textbf{Val} &
     \textbf{Test} &
     \multicolumn{1}{l|}{\textbf{ACC}} &
     \multicolumn{1}{l|}{P} &
     F &
     \multicolumn{1}{l|}{\textbf{ACC}} &
     \multicolumn{1}{l|}{P} &
     F &
     \multicolumn{1}{l|}{\textbf{ACC}} &
     \multicolumn{1}{l|}{P} &
     \multicolumn{1}{l|}{F} &
     \multicolumn{1}{r}{AUC} \\ \hline
    \multicolumn{1}{l|}{a)Semantics/CLIP Text-Image} &
     \multicolumn{1}{l|}{453,128} &
     \multicolumn{1}{l|}{47,248} &
     47,288 &
     \multicolumn{1}{l|}{\textbf{0.68}} &
     \multicolumn{1}{l|}{0.74} &
     0.61 &
     \multicolumn{1}{l|}{\textbf{0.60}} &
     \multicolumn{1}{l|}{0.59} &
     0.61 &
     \multicolumn{1}{l|}{\textbf{0.77}} &
     \multicolumn{1}{l|}{0.76} &
     \multicolumn{1}{l|}{0.79} &
     \multicolumn{1}{r}{0.77} \\ \hline
    \multicolumn{1}{l|}{b)Semantics/CLIP Text-Text} &
     \multicolumn{1}{l|}{516,072} &
     \multicolumn{1}{l|}{53,876} &
     54,164 &
     \multicolumn{1}{l|}{\textbf{0.72}} &
     \multicolumn{1}{l|}{0.74} &
     0.70 &
     \multicolumn{1}{l|}{\textbf{0.62}} &
     \multicolumn{1}{l|}{0.60} &
     0.63 &
     \multicolumn{1}{l|}{\textbf{0.80}} &
     \multicolumn{1}{l|}{0.80} &
     \multicolumn{1}{l|}{0.81} &
     \multicolumn{1}{r}{0.80} \\ \hline
    \multicolumn{1}{l|}{c)Person/SBERT-WK Text-Text} &
     \multicolumn{1}{l|}{17,768} &
     \multicolumn{1}{l|}{1,756} &
     1,816 &
     \multicolumn{1}{l|}{\textbf{0.63}} &
     \multicolumn{1}{l|}{0.70} &
     0.57 &
     \multicolumn{1}{l|}{\textbf{0.55}} &
     \multicolumn{1}{l|}{0.54} &
     0.56 &
     \multicolumn{1}{l|}{\textbf{0.71}} &
     \multicolumn{1}{l|}{0.66} &
     \multicolumn{1}{l|}{0.74} &
     \multicolumn{1}{r}{0.70} \\ \hline
    \multicolumn{1}{l|}{d)Scene/ResNet Place} &
     \multicolumn{1}{l|}{124,860} &
     \multicolumn{1}{l|}{13,588} &
     13,636 &
     \multicolumn{1}{l|}{\textbf{0.71}} &
     \multicolumn{1}{l|}{0.77} &
     0.65 &
     \multicolumn{1}{l|}{\textbf{0.65}} &
     \multicolumn{1}{l|}{0.63} &
     0.67 &
     \multicolumn{1}{l|}{\textbf{0.84}} &
     \multicolumn{1}{l|}{0.83} &
     \multicolumn{1}{l|}{0.85} &
     \multicolumn{1}{r}{0.83} \\ \hline
    \multicolumn{1}{l|}{Merged/Balanced} &
     \multicolumn{1}{l|}{71,072} &
     \multicolumn{1}{l|}{7,024} &
     7,264 &
     \multicolumn{1}{l|}{\textbf{0.65}} &
     \multicolumn{1}{l|}{0.67} &
     0.64 &
     \multicolumn{1}{l|}{\textbf{0.63}} &
     \multicolumn{1}{l|}{0.62} &
     0.64 &
     \multicolumn{1}{l|}{\textbf{0.80}} &
     \multicolumn{1}{l|}{0.78} &
     \multicolumn{1}{l|}{0.81} &
     \multicolumn{1}{r}{0.79} \\ \hline
    \end{tabular}
    }
\end{table}
\begin{table}[t]
     \caption{Comparison results of our proposed approach with other approaches using the same dataset (The merged/balanced split of the NewsCLIPpings dataset) (Accuracy in \%). (NN: Neural Network, LLM: Large Language Model, VLM: Vision-Language Model).}
     \label{tab:comparison_result_other_methods}
     \center
     \resizebox{9cm}{!}{%
     \begin{tabular}{@{}l|l|l|lr@{}}
     \toprule
     \multicolumn{1}{l|}{\textbf{Paper}} &
     \multicolumn{1}{l|}{\textbf{Year}} &
     \multicolumn{1}{l|}{\textbf{Model}} &
     \multicolumn{1}{l|}{\textbf{Detector Based on}} &
     \textbf{ACC} \\ \midrule
     \multicolumn{1}{l|}{Luo \textit{et al.} \cite{luo2021newsclippings}} &
     \multicolumn{1}{l|}{2021} &
     \multicolumn{1}{l|}{CLIP, VisualBERT} &
     \multicolumn{1}{l|}{VLM} &
     65.9 \\ \midrule
     \multicolumn{1}{l|}{Huang \textit{et al.}\cite{huang2022text}} &
     \multicolumn{1}{l|}{2022} &
     \multicolumn{1}{l|}{CLIP, VinVL} &
     \multicolumn{1}{l|}{VLM} &
     65.2 \\ \midrule
     \multicolumn{1}{l|}{Sahar \textit{et al.} \cite{abdelnabi2022open}} &
     \multicolumn{1}{l|}{2022} &
     \multicolumn{1}{l|}{ResNet152, CLIP, SBERT} &
     \multicolumn{1}{l|}{NN, LLM, VLM} &
     66.1 \\ \midrule
     \multicolumn{1}{l|}{Zhang \textit{et al.} \cite{zhang2023detecting}} &
     \multicolumn{1}{l|}{2023} &
     \multicolumn{1}{l|}{\begin{tabular}[c]{@{}l@{}}VisualBERT, CLIP, VinVL,\\FaceNet+BERT\end{tabular}} &
     \multicolumn{1}{l|}{NN, LLM, VLM} &
     62.8 \\ \midrule
     \multicolumn{1}{l|}{Fatma \textit{et al.}~\cite{shalabi2023imagetext}} &
     \multicolumn{1}{l|}{2023} &
     \multicolumn{1}{l|}{CLIP, SBERT, ViT} &
     \multicolumn{1}{l|}{NN, LLM, VLM} &
     68.8 \\ \midrule
     \multicolumn{1}{l|}{\textbf{Ours}} &
     \multicolumn{1}{l|}{2023} &
     \multicolumn{1}{l|}{MiniGPT-4} &
     \multicolumn{1}{l|}{VLM} &
     \textbf{80.0} \\ \bottomrule
     \end{tabular}
     }
\end{table}
\section{Conclusion}
Large vision-language models (LVLMs)  have demonstrated the potential for effective out-of-context (OOC) detection tasks. While initial studies suggest that LVLMs may be unsuitable for OOC detection due to their tendency to provide descriptive responses rather than direct answers, fine-tuning LVLMs on OOC data has shown promising results in improving their OOC detection accuracy. This highlights the importance of fine-tuning LVLMs for specific tasks to enhance their performance. However, further research is needed to address the limitations of LVLMs in providing explanations for their answers. This opacity presents obstacles in comprehending their decision-making process and hinders trust and reliability. To address this gap, we will strive to develop a more robust and interpretable framework for OOC detection by addressing these limitations.

\section*{Acknowledgments}
This work was partially supported by JSPS KAKENHI Grant JP21H04907, and by JST CREST Grants JPMJCR18A6 and JPMJCR20D3, Japan.

%
%
%

\end{document}